\documentclass{article}

\usepackage{jmlr2e}

\usepackage[utf8]{inputenc} 
\usepackage[T1]{fontenc}    
\usepackage{hyperref}       
\usepackage{url}            
\usepackage{booktabs}       
\usepackage{amsfonts}       
\usepackage{nicefrac}       
\usepackage{microtype}      

\usepackage{amsmath}
\usepackage{times}
\usepackage{amsfonts,psfrag,epsfig}
\usepackage{color}
\usepackage{amsmath,amssymb,hyperref}
\usepackage{tabularx}
\usepackage{algorithm}
\usepackage{tikz}
\usetikzlibrary{intersections}
\usetikzlibrary{shapes,arrows}
\usepackage{amsthm}
\usepackage{algorithm}
\usepackage{algpseudocode}
\usepackage{bbm}
\usepackage{float}
\usepackage{relsize}
\usepackage{url}
\usepackage{graphicx}
\usepackage{caption}
\usepackage{subcaption}

\newtheorem{definition}{Definition}
\newtheorem{theorem}{Theorem}
\newtheorem{lemma}{Lemma}
\newtheorem{corollary}{Corollary}

\title{Synthesis of MCMC and Belief Propagation}

\begin{document}
\author{\name Sungsoo Ahn 
\email sungsoo.ahn@kaist.ac.kr 
\AND 
\name Michael Chertkov 
\email chertkov@lanl.gov
\AND 
\name Jinwoo Shin
\email jinwoos@kaist.ac.kr
}
\maketitle
\begin{abstract}
Markov Chain Monte Carlo (MCMC) and Belief Propagation (BP) are the most popular algorithms for computational inference in Graphical Models (GM). 
In principle, MCMC is an exact probabilistic method which, however, often suffers from exponentially slow mixing. In contrast, BP 
is a deterministic method, which is typically fast, 
empirically very successful, however in general lacking control of accuracy over loopy graphs. 
In this paper, we introduce MCMC algorithms correcting the approximation error of BP, i.e.,  we provide a way to compensate for BP errors via a consecutive BP-aware MCMC. 
Our framework is based on
the Loop Calculus approach 
which allows to express the BP error 
as a sum of weighted generalized loops. Although the full series is computationally intractable, 
it is known that a truncated series, summing up all 2-regular loops, is computable in polynomial-time for planar pair-wise binary GMs
and it also provides a highly accurate approximation empirically. Motivated by this, we first propose a polynomial-time
approximation MCMC scheme for the truncated series of general (non-planar) pair-wise binary models. 
Our main idea here is to use the Worm algorithm, known to provide fast mixing in other (related) problems, and then 
design an appropriate rejection scheme to sample
2-regular loops. Furthermore, we also design an efficient rejection-free MCMC scheme 
for approximating the full series. 
The main novelty underlying our design 
is in utilizing the concept of cycle basis, which provides an efficient decomposition of the generalized loops.
In essence, the proposed MCMC schemes run on transformed GM built upon 
the non-trivial BP solution, and
our experiments show that this synthesis of BP and MCMC 
outperforms both direct MCMC and bare BP schemes. 
\end{abstract}

\vspace{-0.1in}
\section{Introduction}
GMs express factorization of the joint multivariate probability distributions in statistics via graph of relations between variables. The concept of GM has been used successfully in information theory, physics, artificial intelligence and machine learning \cite{C1_1,C1_2,C1_3,C1_4,C1_5, C1_6}.  
Of many inference problems one can set with a GM, computing partition function (normalization), or equivalently marginalizing the joint distribution, is the most general problem of interest. However, this paradigmatic inference problem is known to be computationally intractable in general, i.e., formally it is \#P-hard even to approximate \cite{C2, C3}.

To address this obstacle, extensive efforts have been made to develop practical approximation methods, among which MCMC- \cite{C5} based and BP- \cite{C6} based algorithms are, arguably, the most popular and practically successful ones. MCMC is exact, i.e., it converges to the correct answer, but its convergence/mixing is, in general, exponential in the system size. On the other hand, message passing implementations of BP typically demonstrate fast convergence, however in general lacking approximation guarantees for GM containing loops. 
Motivated by this complementarity of the MCMC and BP approaches, we aim here to synthesize a hybrid approach benefiting from a joint use of MCMC and BP. 

At a high level, our proposed scheme uses BP as the first step and then runs MCMC to correct for the approximation error of BP. To design such
an ``error-correcting" MCMC, we utilize the Loop Calculus approach \cite{C8} 
which allows, in a nutshell, 
to express the BP error
as a sum (i.e., series) of weights of the so-called generalized loops (sub-graphs of a special structure). 
There are several challenges one needs to overcome. 
First of all, to design an efficient Markov Chain (MC) sampler, one needs to design a scheme which allows efficient transitions between the generalized loops. Second, even if one designs such a MC which is capable of accessing all the generalized loops, it may mix slowly. 
Finally, weights of generalized loops can be positive or negative, while an individual MCMC can only generate non-negative contributions. 

Since approximating the full loop series (LS) is intractable in general,
we first explore whether we can deal with the challenges at least   in the case of the truncated LS
corresponding to 2-regular loops. In fact,  this problem has been analyzed in the case 
of the planar pairwise binary GMs \cite{C9,C9-1} where it was shown that
the 2-regular LS
is computable 
exactly in polynomial-time through a reduction to a Pfaffian
(or determinant) computation \cite{C10}. 
In particular, 
the partition function of the Ising model
without external field (i.e., where only pair-wise factors present) is computable exactly via 
the 2-regular LS.
Furthermore, the authors show that in the case of general planar pairwise binary GMs, the 2-regular LS
provides a highly accurate approximation empirically.
Motivated by these results,
we address the same question in the general (i.e., non-planar) case of pairwise binary GMs via MCMC. 
For the choice of MC, we adopt
the Worm algorithm \cite{C12}. 
We prove that with some modification including rejections, 
the algorithm allows to sample (with probabilities proportional to respective weights) 2-regular loops in polynomial-time.
Then, we design a novel simulated annealing strategy using the sampler
to estimate separately positive and negative parts of the 2-regular LS.
Given any $\varepsilon>0$, this leads to a $\varepsilon$-approximation polynomial-time scheme for the 2-regular LS
under a mild assumption.

We next turn to estimating the full LS.
In this part, we ignore the theoretical question of establishing the polynomial
mixing time of a MC, and instead focus on designing an empirically efficient MCMC scheme. 
We design an MC using a cycle basis of the graph \cite{C11} to sample generalized loops directly, without rejections. It transits from one generalized loop to another 
by adding or deleting a random element of the cycle basis. 
Using the MC sampler, 
we design a simulated annealing strategy
for estimating the full LS, which is
similar to what was used earlier 
to estimate the 2-regular LS. 
Notice that even though the prime focus of this paper is on pairwise binary GMs, the proposed MCMC scheme
allows straightforward generalization to general non-binary GMs. 

In summary, we propose novel MCMC schemes to estimate the LS correction to the BP contribution to the partition function. 
Since already the bare BP provides a highly non-trivial estimation for the partition function, 
it is naturally expected and confirmed in our experimental results that the proposed algorithm outperforms other standard (not related to BP) MCMC schemes applied to the original GM.
We believe that our approach provides a new angle for
approximate inference on GM and
is of broader interest to various applications involving GMs.

\vspace{-0.05in}
\section{Preliminaries}\label{sec:pre}
\subsection{Graphical models and belief propagation} 
\label{subsec:GM}

Given undirected graph $G=(V,E)$ with $|V|=n, |E|=m$,
a pairwise binary {\it Markov Random Fields} (MRF) defines
the following joint probability distribution on $x=[x_{v} \in\{0,1\}: v \in V]$:
\begin{equation*}
p(x) = \frac{1}{Z} \prod_{v\in V} \psi_{v}(x_{v}) \prod_{(u,v)\in E} \psi_{u,v}(x_{u},x_{v}),
\qquad Z:= \sum_{x\in \{0,1\}^{n}}\prod_{v\in V}\psi_{v}(x_v)\prod_{(u,v)\in E}\psi_{u,v,}(x_{u},x_{v})
\end{equation*}
where $\psi_v,\psi_{u,v}$ are some non-negative functions, called 
{\it compatibility} or {\it factor} functions, and the normalization constant $Z$
is called the {\it partition function}.
Without loss of generality, we assume $G$ is connected. 
It is known that approximating the partition function is \#P-hard in general \cite{C3}.
Belief Propagation (BP) is a popular message-passing heuristic for approximating
marginal distributions of MRF. The BP algorithm  iterates the following message updates for all $(u,v)\in E$:
\begin{equation*}
m^{t+1}_{u\rightarrow v}(x_{v}) 
\propto 
\sum_{x_{u}\in \{0,1\}}
\psi_{u,v}(x_{u},x_{v})\psi_{u}(x_{u})
\prod_{w\in N(u)\backslash v}
m^{t}_{w\rightarrow u}(x_{u}),
\end{equation*}
where 
${N}(v)$ denotes the set of neighbors of $v$.
In general BP may fail to converge, however in this case one may substitute it with a somehow more involved algorithm provably convergent to its fixed point \cite{ConvBP1,ConvBP2,ConvBP3}.
Estimates for the marginal probabilities are expressed via the fixed-point messages $\{m_{u\to v}:(u,v)\in E\}$ as follows: $\tau_{v}(x_{v}) \propto \psi_{v}(x_{v})
\prod_{u \in N(v)}m_{u\rightarrow v}(x_{v})$ and
\begin{align*}
\tau_{u,v}(x_{u},x_{v}) &\propto \psi_{u}(x_{u})\psi_{v}(x_{v})\psi_{u,v}(x_{u},x_{v})
\left(
\prod_{w \in {N}(u)}m_{w\rightarrow v}(x_{u})
\right)
\left(
\prod_{w \in {N}(v)}m_{w\rightarrow v}(x_{v})
\right).
\end{align*}

\vspace{-0.05in}
\subsection{Bethe approximation and loop calculus}

BP marginals also results in the following {\it Bethe approximation} for the partition function $Z$:  
\begin{align*}
\begin{split}
\log Z_{\text{Bethe}} 
= & \sum_{v \in V}\sum_{x_{v}} \tau_{v}(x_{v}) \log \psi_{v}(x_{v}) 
+ \sum_{(u,v) \in E}\sum_{x_{u}, x_{v}} \tau_{u,v}(x_{u},x_{v}) \log \psi_{u,v}(x_{u},x_{v}) \\
- & \sum_{v \in V}\sum_{x_{v}} \tau_{v}(x_{v}) \log \tau_{v}(x_{v}) 
- \sum_{(u,v) \in E}\sum_{x_{u}, x_{v}} \tau_{u,v}(x_{u},x_{v}) \log \frac{\tau_{u,v}(x_{u},x_{v})}{\tau_{u}(x_{u})\tau_{v}(x_{v})}
\end{split}
\end{align*}
If graph $G$ is a tree,
the Bethe approximation is exact, i.e., $Z_{\text{Bethe}} = Z$. 
However, in general, i.e. for the graph with cycles, BP algorithm provides often rather accurate but still an approximation. 


{\it Loop Series} (LS) \cite{C8} expresses, $Z/Z_{\text{Bethe}}$, as 
the following sum/series:
\begin{equation*}
\frac{Z}{Z_{\text{Bethe}}} ~=~  Z_{\text{Loop}}
~:=~\sum_{F\in \mathcal L} w(F),\quad w(\emptyset)=1,
\end{equation*}
\begin{equation*}
w(F) := 
\prod_{(u,v) \in E_{F}}\left(\frac{\tau_{u,v}(1,1)}{\tau_{u}(1)\tau_{v}(1)}-1\right) 
\prod_{v\in V_{F}}\left(\tau_{v}(1)+(-1)^{d_{F}(v)}\left(\frac{\tau_{v}(1)}{1-\tau_{v}(1)}\right)^{d_{F}(v)-1}\tau_{v}(1)\right) \label{eq:weight}
\end{equation*}
where each term/weight is associated with the so-called {\it generalized loop} $F$ 
and
$\mathcal{L}$ denotes the set of all generalized loops in graph $G$ (including the empty subgraph $\emptyset$).
Here, a subgraph $F$ of $G$ is called generalized loop
if all vertices $v\in F$ have degree $d_{F}(v)$ (in the subgraph) no smaller than $2$. 

Since the number of generalized loops is exponentially large, computing $Z_{\text{Loop}}$
is intractable in general. However, 
the following truncated sum of ${Z_{\text{Loop}}}$, called {\it 2-regular loop series},
is known to be computable in polynomial-time if $G$ is planar \cite{C9}:\footnote{
Note that the number of 2-regular loops is exponentially large in general.}
$$Z_{\text{2-Loop}} := \sum_{F \in \mathcal{L}_{\text{2-Loop}}} w(F),$$
where $\mathcal{L}_{\text{2-Loop}}$ denotes the set of all 
{\it 2-regular generalized loops}, i.e., $F\in 
\mathcal{L}_{\text{2-Loop}}$ if $d_F(v)=2$ for every vertex $v$ of $F$.
One can check that $Z_{\text{Loop}}=Z_{\text{2-Loop}}$ for 
the Ising model without the external fields. Furthermore, as stated in \cite{C9,C9-1} for the general case, 
$Z_{\text{2-Loop}}$ provides a good empirical estimation for $Z_{\text{Loop}}$.

\vspace{-0.05in}
\section{Estimating 2-regular loop series via MCMC}\label{sec:2regular}

In this section, we aim to describe how the $2$-regular loop series  $Z_{\text{2-Loop}}$ can be estimated in polynomial-time. To this end, we first assume that the maximum degree $\Delta$ of the graph $G$ is at most 3. 
This degree constrained assumption is not really restrictive
since any pairwise binary model can be easily expressed as an equivalent one with $\Delta\leq 3$, e.g.,
see the supplementary material.
The rest of this section consists of two parts. 
We first propose an algorithm generating
a 2-regular loop sample with the probability proportional to the absolute value of 
its weight, i.e.,
\begin{equation*}
\pi_{\text{2-Loop}}(F) := \frac{|w(F)|}{Z^{\dagger}_{\text{2-Loop}}},\qquad
\mbox{where}~~ Z^{\dagger}_{\text{2-Loop}}  = \sum_{F \in \mathcal{L}_{\text{2-Loop}}} |w(F)|.
\end{equation*}
Note that this 2-regular loop contribution allows the following factorization: for any $F\in \mathcal{L}_{\text{2-Loop}}$,
\begin{equation}\label{eq:edgeweight}
|w(F)| = \prod_{e\in F} w(e),\qquad\mbox{where}\quad
w(e):=\left| \frac{\tau_{u,v}(1,1)-\tau_{u}(1)\tau_{v}(1)}{\sqrt{\tau_{u}(1)\tau_{v}(1)(1-\tau_{u}(1))(1-\tau_{v}(1))}}\right|.
\end{equation}
In the second part we use the sampler constructed in the first part to design a simulated annealing scheme to estimate $Z_{\text{2-Loop}}$.

\vspace{-0.05in}
\subsection{Sampling 2-regular loops}\label{sec:sample2regular}

We suggest to sample the 2-regular loops distributed according to $\pi_{\text{2-Loop}}$ through a version of the Worm algorithm proposed by Prokofiev and Svistunov \cite{C12}.
It can be viewed as a MC exploring the set, $\mathcal L_{\text{2-Loop}}\bigcup \mathcal L_{\text{2-Odd}}$, where
$\mathcal L_{\text{2-Odd}}$ is the set of all subgraphs of $G$ with
exactly two odd-degree vertices.
Given current state $F\in \mathcal L_{\text{2-Loop}}\bigcup \mathcal L_{\text{2-Odd}}$,
it chooses the next state $F^\prime$ as follows:

\begin{enumerate}
\item If $F\in \mathcal L_{\text{2-Odd}}$, pick a random vertex $v$ (uniformly) from  $V$. Otherwise, 
pick a random odd-degree vertex $v$ (uniformly) from $F$.
\item Choose a random neighbor $u$ of $v$ (uniformly) within $G$, and set $F^\prime \leftarrow F$ initially.
\item Update $F^\prime \leftarrow F \oplus \{u,v\}$ with the probability
\begin{equation*}
\begin{cases}
\min \left(\frac{1}{n} \frac{|w(F\oplus \{u,v\})|}{|w(F)|}, 1\right)&\qquad \mbox{if}~ F\in \mathcal{L}_{\text{2-Loop}}\\
\min \left(\frac{n}{4} \frac{|w(F\oplus \{u,v\})|}{|w(F)|}, 1\right)&\qquad \mbox{else if}~ F \oplus \{u,v\}\in \mathcal{L}_{\text{2-Loop}}\\
\min \left(\frac{d(v)}{2d(u)} \frac{|w(F\oplus \{u,v\})|}{|w(F)|}, 1\right) &\qquad \mbox{else if}~ F, F \oplus \{u,v\} \in \mathcal{L}_{\text{2-Odd}}
\end{cases}\end{equation*}
\end{enumerate}
Here, $\oplus$ denotes the symmetric difference and
for $F\in \mathcal L_{\text{2-Odd}}$, its weight is defined according to
$w(F) = \prod_{e\in F} w(e)$. In essence, the Worm algorithm consists in  either deleting or adding an edge to the current subgraph $F$. 
From the Worm algorithm, we transition to the following algorithm which samples $2$-regular loops with probability $\pi_{\text{2-Loop}}$ simply by adding rejection of $F$ if $F\in \mathcal L_{\text{2-Odd}}$.

\begin{algorithm}[H]
\caption{Sampling 2-regular loops}
\label{alg:sample2regular}
\begin{algorithmic}[1]
\State {\bf Input:} Number of trials $N$; number of iterations $T$ of the Worm algorithm 
\State {\bf Output:} 2-regular loop $F$.
\For{$i = 1 \to N$} 
\State Set $F\leftarrow \emptyset$ and update it
$T$ times by running the Worm algorithm 
\If{$F$ is a 2-regular loop} 
\State BREAK and output $F$.
\EndIf
\EndFor
\State Output $F=\emptyset$. 
\end{algorithmic}
\end{algorithm}
The following theorem states that {Algorithm \ref{alg:sample2regular}} can generate a desired 
random sample in polynomial-time.
\begin{theorem}\label{thm:sample2regular}
Given $\delta>0$, choose 
inputs of {Algorithm \ref{alg:sample2regular}} as 
\begin{equation*}
N \geq 1.2\, n\log(3\delta^{-1}),\qquad\mbox{and}\qquad
T \geq 
(m-n+1)\log 2 + 4\Delta mn^{4}\log (3n\delta^{-1}).
\end{equation*}
Then, it follows that
\begin{equation*}\frac12\sum_{F\in \mathcal L_{\text{2-Loop}}}
\left|P\bigg[\mbox{{Algorithm \ref{alg:sample2regular}} outputs $F$}\bigg] - \pi_{\text{2-Loop}}(F) \right| \leq \delta.
\end{equation*}
namely, the total variation distance between $\pi_{\text{2-Loop}}$ and the output distribution of 
{Algorithm \ref{alg:sample2regular}} is at most $\delta$.
\end{theorem}
The proof of the above theorem is presented in the supplementary material due to the space constraint.
In the proof,
we first show that MC induced by the Worm algorithm mixes in polynomial time, and then prove that acceptance of a 2-regular loop, i.e., line 6 of {Algorithm \ref{alg:sample2regular}}, occurs with high probability.
Notice that the uniform-weight version of the former proof, i.e., fast mixing, was recently proven in \cite{C13}. For completeness of the material exposition, we present the general case proof of interest for us. 
The latter proof, i.e., high acceptance, requires to bound $|\mathcal L_{\text{2-Loop}}|$
and $|\mathcal L_{\text{2-Odd}}|$ to show that the probability of sampling 2-regular loops under 
the Worm algorithm 
is $1/\text{poly}(n)$ for some polynomial function $\text{poly}(n)$.

\vspace{-0.05in}
\subsection{Simulated annealing for approximating 2-regular loop series}
\label{sec:2regularsec2}

Here we utilize Theorem \ref{thm:sample2regular} to describe an algorithm approximating the $2$-regular LS $Z_{\text{2-Loop}}$ in polynomial time. To achieve this goal, 
we rely on the simulated annealing strategy \cite{SA} which requires to decide
a monotone cooling schedule  $\beta_0,\beta_1,\dots, \beta_{\ell-1},\beta_\ell$, where $\beta_\ell$ corresponds to the target counting problem and $\beta_0$ does to its relaxed easy version. Thus, designing
an appropriate cooling strategy 
is the first challenge to address. We will also describe how to deal with the issue that $Z_{\text{2-Loop}}$ is
a sum of positive and negative terms, while most simulated annealing strategies in the literature mainly studied on sums of non-negative terms. 
This second challenge is related to the so-called `fermion sign problem' common in statistical mechanics of quantum systems
\cite{sign}. 
Before we describe the proposed algorithm in details, let us provide its intuitive sketch.

The proposed algorithm consists of two parts: a) estimating $Z^{\dagger}_{\text{2-Loop}}$ via a simulated annealing strategy and b) estimating $Z_{\text{2-Loop}}/Z^{\dagger}_{\text{2-Loop}}$ via counting samples corresponding to negative terms in the 2-regular loop series.
First consider the following $\beta$-parametrized, auxiliary distribution over $2$-regular loops: 
\begin{align}
\label{pi-2-Loop}
\pi_{\text{2-Loop}}
(F:\beta) = \frac{1}{Z^{\dagger}_{\text{2-Loop}}(\beta)}|w(F)|^{\beta},\qquad
\mbox{for}~~ 0\leq \beta\leq 1.
\end{align}
Note that one can generate samples approximately with probability (\ref{pi-2-Loop})
in polynomial-time using {Algorithm \ref{alg:sample2regular}} by setting $w \leftarrow w^{\beta}$.
Indeed, it follows that for $\beta^\prime> \beta$, 
\begin{equation*}
\frac{Z^{\dagger}_{\text{2-Loop}}(\beta^\prime)}{Z^{\dagger}_{\text{2-Loop}}(\beta)}
= 
\sum_{F \in \mathcal{L}_{\text{2-Loop}}}
|w(F)|^{\beta^\prime - \beta}
\frac{|w(F)|^{\beta}}{Z^{\dagger}_{\text{2-Loop}}(\beta)}
=
\mathbb{E}_{\pi_{\text{2-Loop}}(\beta)}\left[|w(F)|^{\beta^\prime - \beta}\right],
\end{equation*}
where the expectation 
can be estimated using $O(1)$ samples if it is $\Theta(1)$, i.e., $\beta^\prime$ is sufficiently close to $\beta$.
Then, for any increasing sequence $\beta_0=0,\beta_1,\dots, \beta_{n-1},\beta_n=1$,   
we derive
\begin{equation*}
Z^{\dagger}_{\text{2-Loop}} = 
\frac{Z^{\dagger}_{\text{2-Loop}}(\beta_n)}{Z^{\dagger}_{\text{2-Loop}}(\beta_{n-1})}\cdot 
\frac{Z^{\dagger}_{\text{2-Loop}}(\beta_{n-1})}{Z^{\dagger}_{\text{2-Loop}}(\beta_{n-2})}\cdots
\frac{Z^{\dagger}_{\text{2-Loop}}(\beta_{2})}{Z^{\dagger}_{\text{2-Loop}}(\beta_{1})}
\frac{Z^{\dagger}_{\text{2-Loop}}(\beta_{1})}{Z^{\dagger}_{\text{2-Loop}}(\beta_0)}
Z^{\dagger}_{\text{2-Loop}}(0),
\end{equation*}
where it is know that 
$Z^{\dagger}_{\text{2-Loop}}(0)$, i.e., the total number of 2-regular loops, is  
exactly $2^{m-n+1}$ \cite{C11}. 
This allows us to estimate $Z^{\dagger}_{\text{2-Loop}}$ simply by estimating 
$\mathbb{E}_{\pi_{\text{2-Loop}}(\beta_i)}\left[|w(F)|^{\beta_{i+1} - \beta_i}\right]$
for all $i$.

Our next step is to estimate the ratio $Z_{\text{2-Loop}}/Z^\dagger_{\text{2-Loop}}$.
Let $\mathcal{L}^{-}_{\text{2-Loop}}$ denote the 
set of negative 2-regular loops, i.e.,
\begin{equation*}
\mathcal{L}^{-}_{\text{2-Loop}} := \{F: F\in\mathcal{L}_{\text{2-Loop}}, w(F) < 0\}.
\end{equation*}
Then, the $2$-regular loop series can be expressed as
\begin{equation*}
Z_{\text{2-Loop}} = \left(1-2\frac{\sum_{F \in \mathcal{L}^{-}_{\text{2-Loop}}}|w(F)|}{Z^\dagger_{\text{2-Loop}}}
\right)Z^\dagger_{\text{2-Loop}}
= \left(1-2 P_{\pi_{\text{2--Loop}}}\bigg[w(F)<0\bigg]
\right)Z^\dagger_{\text{2-Loop}},
\end{equation*}
where we estimate $P_{\pi_{\text{2--Loop}}}\big[w(F)<0\big]$ again using samples generated by
{Algorithm \ref{alg:sample2regular}}. 

We provide the formal description of the proposed algorithm and its error bound as follows.
\begin{algorithm}[H]
\caption{Approximation for $Z_{\text{2-Loop}}$}
\label{alg:estimate2regular}
\begin{algorithmic}[1]
\State {\bf Input:}  
Increasing sequence $\beta_0=0<\beta_1<\dots< \beta_{n-1}<\beta_n=1$; number of samples $s_{1}, s_{2}$; number of trials $N_{1}$;  
number of iterations $T_{1}$ for {Algorithm \ref{alg:sample2regular}}.
\For{$i = 0 \to n-1$}
\State Generate 2-regular loops $F_1,\dots, F_{s_1}$ for $\pi_{\text{2-Loop}}(\beta_{i})$ using {Algorithm \ref{alg:sample2regular}} with input $N_{1}$ and $T_{1}$, and set
\begin{equation*}
H_{i} \leftarrow \frac1{s_1}\sum_j w(F_j)^{\beta_{i+1}-\beta_i}.
\end{equation*}
\EndFor
\State Generate 2-regular loops $F_1,\dots, F_{s_2}$ for $\pi_{\text{2-Loop}}$ using {Algorithm \ref{alg:sample2regular}} with input $N_{2}$ and $T_{2}$, and set
\begin{equation*}
\kappa \leftarrow \frac{|\{F_j: w(F_j)<0\}|}{s_2}.
\end{equation*}
\State {\bf Output:} $\widehat Z_{\text{2-Loop}}\leftarrow
(1-2\kappa)2^{m-n+1}\prod_{i}H_{i}$.
\end{algorithmic}
\end{algorithm}

\begin{theorem}\label{thm:estimatepartition2regular}
Given $\varepsilon, \nu > 0$, choose 
inputs of {Algorithm \ref{alg:estimate2regular}} as $\beta_{i} = i/n$ for $i=1,2,\dots,n-1$,
\begin{align*} 
&s_{1}\geq 18144 n^{2}\varepsilon^{-2}w^{-1}_{\min}\lceil\log (6n\nu^{-1})\rceil,\qquad
&&N_{1} \geq  1.2n\log(144n \varepsilon^{-1}w_{\min}^{-1}),\\
&T_{1} \geq
(m-n+1)\log 2 + 4\Delta mn^{4}\log(48 n \varepsilon^{-1}w_{\min}^{-1}),\\
&s_{2} \geq  18144\zeta(1-2\zeta)^{-2}
\varepsilon^{-2}\lceil\log (3\nu^{-1})\rceil,
\qquad\qquad\qquad
&&N_{2} \geq
1.2n\log (144\varepsilon^{-1}
(1-2\zeta)^{-1}),\\
&T_{2} \geq 
(m-n+1)\log 2 + 4\Delta mn^{4}\log(48\varepsilon^{-1}
(1-2\zeta)^{-1})
\end{align*}
where $w_{\min}=\min_{e\in E} w(e)$ and $\zeta = P_{\pi_{\text{2--Loop}}}[w(F)<0].$
Then, the following statement holds
\begin{align*}
    P\left[\frac{|\widehat{Z}_{\text{2-Loop}} - Z_{\text{2-Loop}}|}{Z_{\text{2-Loop}}}\leq 
    \varepsilon
    \right] \leq 1-\nu,
\end{align*}
which means {Algorithm \ref{alg:estimate2regular}} estimates 
$Z_{\text{2-Loop}}$ 
within approximation ratio $1\pm\varepsilon$ with high probability.
\end{theorem}

The proof of the above theorem is presented in the supplementary material due to the space constraint.
We note that all constants entering in Theorem \ref{thm:estimatepartition2regular} were not optimized.
Theorem \ref{thm:estimatepartition2regular} implies that complexity of {Algorithm \ref{alg:estimate2regular}} is polynomial with respect to
$n, 1/\varepsilon, 1/\nu$ under assumption that
$w_{\min}^{-1}$ and $1 - 2P_{\pi_{\text{2--Loop}}}[w(F)<0]$
are polynomially small.  Both $w_{\min}^{-1}$ and $1 - 2P_{\pi_{\text{2--Loop}}}[w(F)<0]$ depend on the choice of BP fixed point, however it is unlikely (unless a degeneracy) that these characteristics become large. In particular, 
$P_{\pi_{\text{2--Loop}}}[w(F)<0]=0$ in the case of attractive models \cite{C15}.

\vspace{-0.05in}
\section{Estimating full loop series via MCMC}\label{sec:fullLS}

In this section, we aim for estimating the full loop series
$Z_{\text{Loop}}$. 
To this end, we design a novel MC sampler for generalized loops, which
adds (or removes) a cycle basis or a path
to (or from) the current generalized loop.
Therefore,  we naturally start this section introducing necessary backgrounds on {\it cycle basis}.
Then, we turn to describe the design of MC sampler for generalized loops. Finally, we describe a simulated annealing scheme similar to the one described in the preceding section. 
We also report its experimental performance comparing with other methods.

\vspace{-0.05in}

\subsection{Sampling generalized loops with cycle basis}\label{sec:sampleGL}

The cycle basis $\mathcal C$ of the graph $G$ is a minimal set of
cycles which allows to represent every 
Eulerian subgraph of $G$  (i.e., subgraphs containing no odd-degree vertex) as a symmetric difference of 
cycles in the set \cite{C11}.
Let us characterize the combinatorial structure of the 
generalized loop using the cycle basis. To this end, consider a set of paths between any pair of vertices:
\begin{align*}
    \mathcal{P} = \{P_{u,v}: u\neq v, u,v\in V, \mbox{$P_{u,v}$ is a path from $u$ to $v$}\},
\end{align*}
i.e., $|\mathcal P| = \binom{n}{2}$. 
Then the following theorem allows to decompose any generalized loop with respect to any selected $\mathcal C$
and $\mathcal P$.
\begin{theorem}\label{thm:generalizedloopdecomp}
Consider any cycle basis $\mathcal C$ and path set $\mathcal P$.
Then, for any generalized loop $F$, there exists a decomposition, $\mathcal{B}\subset \mathcal{C}\cup \mathcal{P}$, such that 
$F$ can be expressed as a symmetric difference of the elements of $\mathcal{B}$, i.e.,
    $F = B_{1} \oplus B_{2} \oplus \cdots B_{k-1} \oplus B_{k}$ for some $B_{i} \in \mathcal{B}$.
\end{theorem}

The proof of the above theorem is given in the supplementary material due to the space constraint. 
Now given any choice of $\mathcal C, \mathcal P$,
consider the following transition from  $F\in \mathcal L$,
to the next state $F^\prime$:
\begin{enumerate}
\item Choose, uniformly at random, an element $B \in\mathcal{C}\cup\mathcal{P}$, and set
$F^\prime\leftarrow F$ initially.
\item If $F\oplus B \in \mathcal{L}$, update 
$F^\prime 
\leftarrow
\begin{cases}
F \oplus B &\mbox{with probability}~
\min \left\{1, \frac{|w(F\oplus B|)|}{|w(F)|}\right\} \\
F&\mbox{otherwise}
\end{cases}.$
\end{enumerate}
Due to Theorem \ref{thm:generalizedloopdecomp},  it is easy to check that
the proposed MC is irreducible and aperiodic, i.e., ergodic, and
the distribution of its $t$-th state converges to the following stationary distribution as $t\to \infty$:
\begin{equation*}
\pi_{\text{Loop}}(F) = \frac{|w(F)|}{Z^{\dagger}_{\text{Loop}}},\qquad
\mbox{where}~~ Z^{\dagger}_{\text{Loop}}  = \sum_{F \in \mathcal{L}_{\text{Loop}}} |w(F)|.
\end{equation*}
One also has a freedom in choosing $\mathcal C, \mathcal P$. To accelerate mixing of MC,
we suggest to choose the minimum weighted cycle basis $\mathcal{C}$ and 
the shortest paths $\mathcal{P}$ 
with respect to the edge weights $\{\log w(e)\}$ defined in \eqref{eq:edgeweight},
which are computable using the algorithm in \cite{C11}
and the Bellman-Ford algorithm \cite{BF}, respectively.
This 
encourages transitions between generalized loops with similar weights.

\vspace{-0.05in}
\subsection{Simulated annealing for approximating full loop series}\label{sec:whole}
\begin{algorithm}
\caption{Approximation for $Z_{\text{Loop}}$}
\label{alg:approximatingfull}
\begin{algorithmic}[1]
\State {\bf Input:} 
Decreasing sequence $\beta_{0} > \beta_{1}> \cdots > \beta_{\ell-1} > \beta_{\ell} = 1$; number of samples $s_{0}$, $s_{1},s_{2}$;
number of iterations $T_{0}$, $T_{1},T_{2}$ for the MC described in Section \ref{sec:sampleGL}
\State 
Generate generalized loops $F_{1}, \cdots, F_{s_{0}}$ by running $T_{0}$ iterations of the MC described in Section \ref{sec:sampleGL} for 
$\pi_{\text{Loop}}(\beta_0)$, and set 
\begin{equation*}
    U \leftarrow \frac{s_{0}}{s^{*}}|w(F^{*})|^{\beta_0},
\end{equation*}
where $F^{*} = \arg\max_{F\in\{F_{1}, \cdots, F_{s_{0}}\}} |w(F)|$ and $s^{*}$ is the number of $F^{*}$ sampled.
\For{$i=0 \to \ell-1$}
\State Generate generalized loops $F_{1}, \cdots, F_{s_{1}}$
by running $T_{1}$ iterations of the MC described in Section \ref{sec:sampleGL} 
 for $\pi_{\text{Loop}}(\beta_i)$,
and set 
    $H_{i}\leftarrow \frac{1}{s_{1}}\sum_{j}|w(F_{j})|^{\beta_{i+1}-\beta_{i}}.$
\EndFor
\State Generate generalized loops $F_{1}, \cdots F_{s_{2}}$
by running $T_2$ iterations of the MC described in Section \ref{sec:sampleGL} for $\pi_{\text{Loop}}$, 
and set
\begin{equation*}
\kappa \leftarrow \frac{|\{F_{j}: w(F_{j}) < 0\}|}{s_{2}}.
\end{equation*}
\State {\bf Output:} $\widehat{Z}_{\text{Loop}} \leftarrow (1-2\kappa)\prod_{i}H_{i}U$.
\end{algorithmic}
\end{algorithm}

Now we are ready to describe a simulated annealing scheme for 
estimating $Z_{\text{Loop}}$. 
It is similar, in principle, with that in   
Section \ref{sec:2regularsec2}. 
First, we again introduce the following $\beta$-parametrized, auxiliary probability distribution
$\pi_{\text{Loop}}(F:\beta) =| w(F)|^{\beta}/Z^{\dagger}_{\text{Loop}}(\beta)$.
For any decreasing sequence of annealing parameters, $\beta_{0}, \beta_{1} , \cdots, \beta_{\ell-1}, \beta_{\ell} = 1$, 
we derive
\begin{equation*}
Z^\dagger_{\text{Loop}} = 
\frac{Z^{\dagger}_{\text{Loop}}(\beta_\ell)}{Z^{\dagger}_{\text{Loop}}(\beta_{\ell-1})}\cdot 
\frac{Z^{\dagger}_{\text{Loop}}(\beta_{\ell-1})}{Z^{\dagger}_{\text{Loop}}(\beta_{\ell-2})}\cdots
\frac{Z^{\dagger}_{\text{Loop}}(\beta_{2})}{Z^{\dagger}_{\text{Loop}}(\beta_{1})}\cdot
\frac{Z^{\dagger}_{\text{Loop}}(\beta_{1})}{Z^{\dagger}_{\text{Loop}}(\beta_0)}
Z^{\dagger}_{\text{Loop}}(\beta_{0}).
\end{equation*}
Following similar procedures in Section \ref{sec:2regularsec2},
one can estimate
${Z^{\dagger}_{\text{Loop}}(\beta^{\prime})}/{Z^{\dagger}_{\text{Loop}}(\beta)} = \mathbb{E}_{\pi_{\text{Loop}}(\beta)}[|w(F)|^{\beta^{
\prime} - \beta}]$
using the sampler described in Section \ref{sec:sampleGL}. 
Moreover, $Z^{\dagger}_{\text{Loop}}(\beta_{0}) = |w(F^{*})|/P_{\pi_{\text{Loop}}(\beta_{0})}(F^{*})$
is estimated by sampling generalized loop $F^{*}$ 
with the highest probability $P_{\pi_{\text{Loop}}(\beta_{0})}(F^{*})$. 
For large enough $\beta_{0}$, 
the approximation error becomes relatively small since  
$P_{\pi_{\text{Loop}}(\beta_{0})}(F^{*}) \propto |w(F^{*})|^{\beta_{0}}$ 
dominates over the distribution.
In combination, this provides a desired approximation for $Z_{\text{Loop}}$.
The result is stated formally in Algorithm \ref{alg:approximatingfull}.

\subsection{Experimental results}\label{sec:exp}
\begin{figure}[t!]
\vspace{-0.1in}
\begin{subfigure}{0.33\textwidth}
  \includegraphics[width=1\linewidth]{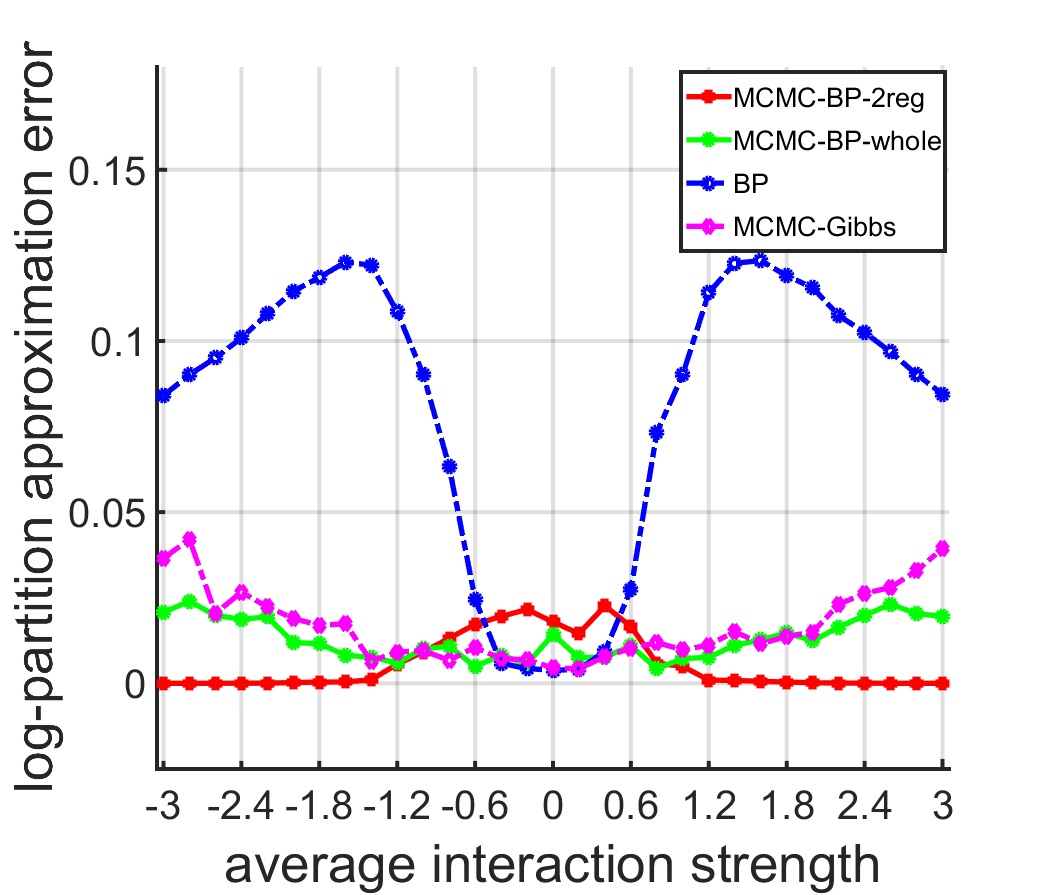}
  \label{fig:sub1}
  \vspace{-15pt}
  \caption{}
\end{subfigure}
\begin{subfigure}{0.33\textwidth}
  \includegraphics[width=1\linewidth]{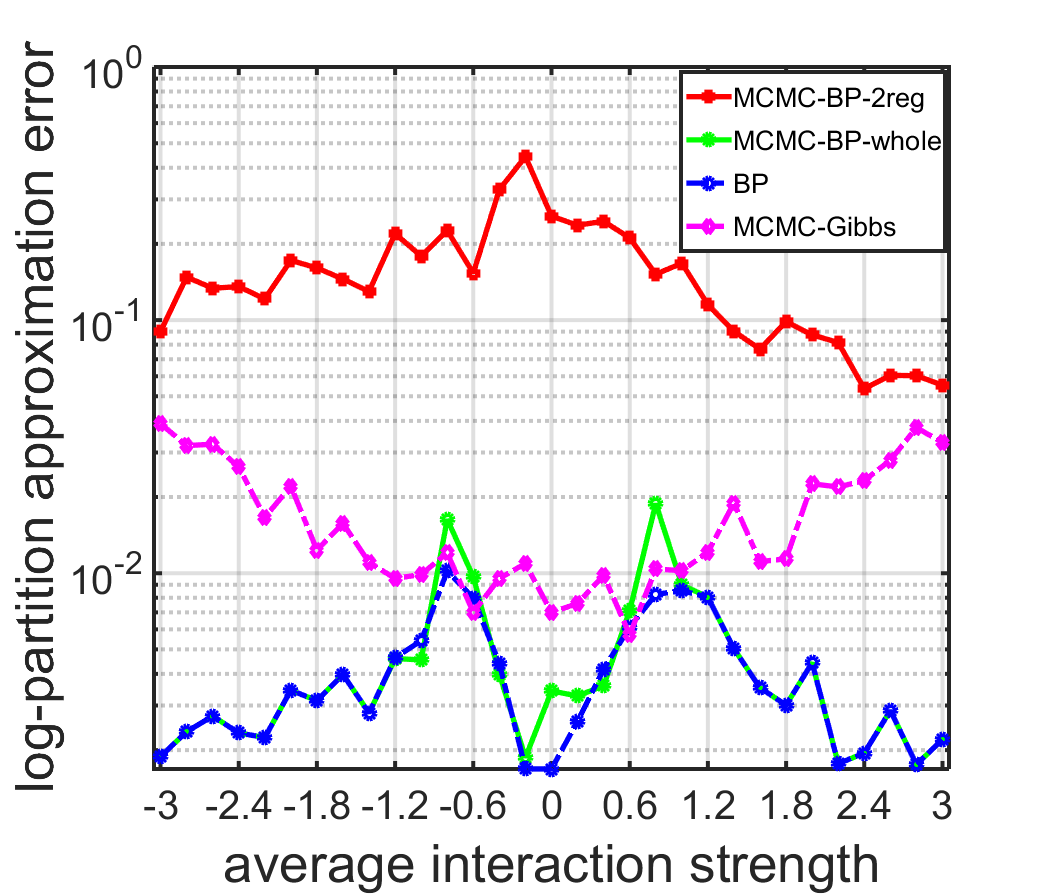}
  \label{fig:sub2}
    \vspace{-15pt}
    \caption{}
\end{subfigure}
\begin{subfigure}{0.33\textwidth}
  \includegraphics[width=1\linewidth]{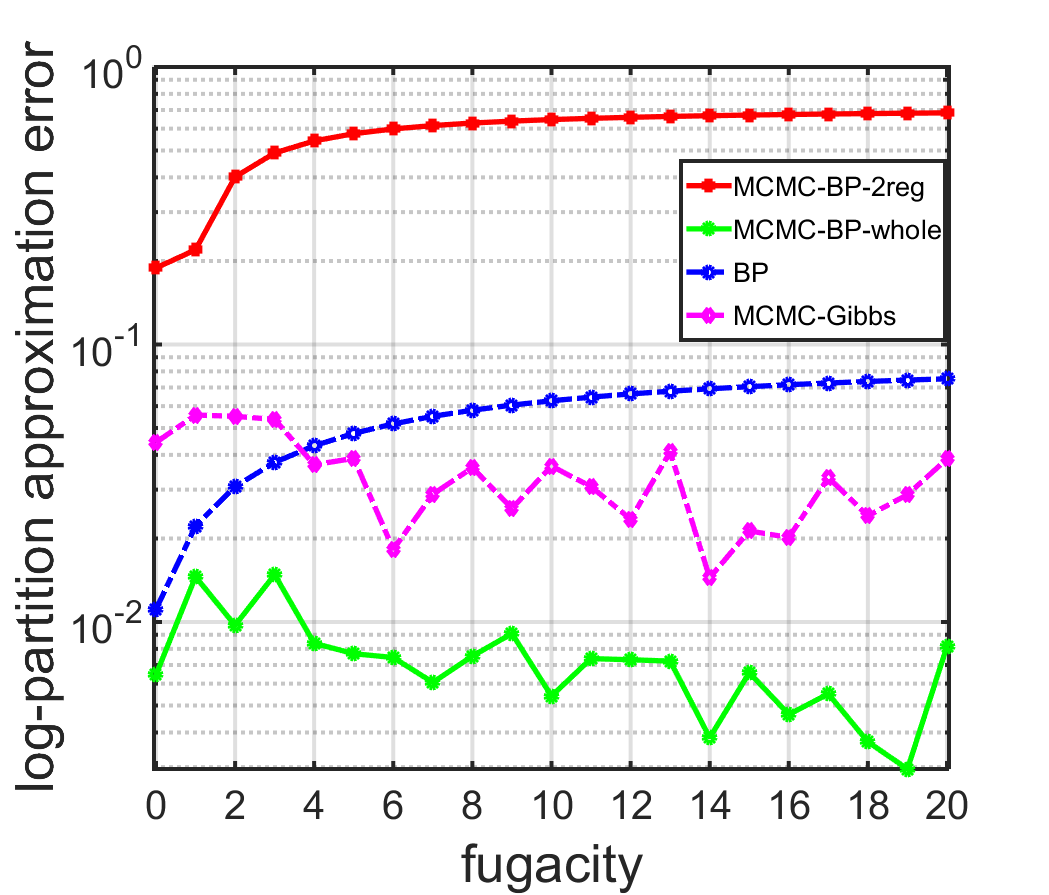}
    \vspace{-15pt}
    \caption{}
\end{subfigure}
\caption{Plots of the 
log-partition function approximation error 
with respect to (average) interaction strength. Each point is averaged over $20$ (random) models. 
}
\label{fig:experiment}
\end{figure}
In this section, we report experimental results for 
computing partition function of the Ising model and the hard-core model. 
We compare 
Algorithm \ref{alg:estimate2regular} in Section \ref{sec:2regular} (coined MCMC-BP-2reg) and 
Algorithm \ref{alg:approximatingfull} in Section \ref{sec:whole} (coined MCMC-BP-whole), 
with 
the bare Bethe approximation (coined BP) 
and the popular Gibbs-sampler (coined MCMC-Gibbs).
To make the comparison fair, we use the same annealing scheme for all MCMC schemes,
thus making their running times comparable.
More specifically, we 
generate each sample after running $T_{1} = 1,000$ iterations of an MC and 
take $s_{1} =100$ samples to compute each estimation (e.g., $H_{i}$) at intermediate steps.
For performance measure, we use the log-partition function 
approximation error defined as $|\log{Z} - \log{Z_{\text{approx}}}|/|\log{Z}|$, 
where $Z_{\text{approx}}$ is the output of the respective algorithm. 
We conducted 3 experiments on the $4\times 4$ grid graph.
In our first experimental setting, we consider the Ising model 
with varying interaction strength and no external (magnetic) field. 
To prepare the model of interest, we start from the Ising model with uniform 
(ferromagnetic/attractive and anti-ferromagnetic/repulsive) 
interaction strength and then  add `glassy' variability in the interaction strength modeled 
via i.i.d Gaussian random variables with mean $0$ and variance $0.5^{2}$, 
i.e. $\mathcal{N}(0,0.5^{2})$. In other words, given average interaction strength $0.3$,
each interaction strength 
in the model is independently chosen as $\mathcal{N}(0.3,0.5^{2})$. 
The second experiment was conducted by adding $\mathcal{N}(0,0.5^2)$ corrections to the external fields 
under the same condition as in the first experiment. 
In this case we observe that
BP often fails to converge, and use the Concave Convex Procedure (CCCP) \cite{ConvBP2} for finding 
BP fixed points.
Finally, we experiment with the hard-core model on the $4\times 4$ grid graph 
with varying a positive parameter $\lambda>0$, called `fugacity' \cite{Independent}. 
As seen clearly in Figure \ref{fig:experiment}, 
BP and MCMC-Gibbs are outperformed by 
MCMC-BP-2reg or MCMC-BP-whole at most tested regimes 
in the first experiment with no external field,
where in this case, the 2-regular loop series (LS) is equal to the full one.
Even in the regimes where MCMC-Gibbs outperforms BP, 
our schemes correct the error of BP and performs at least as good as MCMC-Gibbs. 
In the experiments, we observe that advantage of our schemes over BP 
is more pronounced when the error of BP is large. 
A theoretical reasoning behind this observation
is as follows. If the performance of BP is good, i.e. the loop series (LS) is close to $1$, the 
contribution of empty generalized loop, i.e., $w(\emptyset)$, in LS is significant, 
and it becomes harder 
to sample other generalized loops accurately. 

\vspace{-0.05in}
\section{Conclusion}

In this paper, we propose new MCMC schemes for approximate inference in GMs.
The main novelty of our approach is in designing BP-aware MCs utilizing the non-trivial BP solutions. 
In experiments, our BP based MCMC scheme also outperforms other alternatives. We anticipate that 
this new technique will be of interest to many applications where GMs are used for statistical reasoning. 
\section*{Acknowledgements.}
The work of MC  was carried out under the auspices of the National Nuclear Security Administration of the U.S. Department of
Energy at Los Alamos National Laboratory under Contract
No. DE-AC52-06NA25396, and was partially supported by the
Advanced Grid Modeling Program in the U.S. Department
of Energy Office of Electricity.

\appendix
\appendix

\section{Transformation to an equivalent binary pairwise model with maximum degree at most 3}
\begin{figure}[H]
    \centering
    \includegraphics[width=0.6\textwidth]{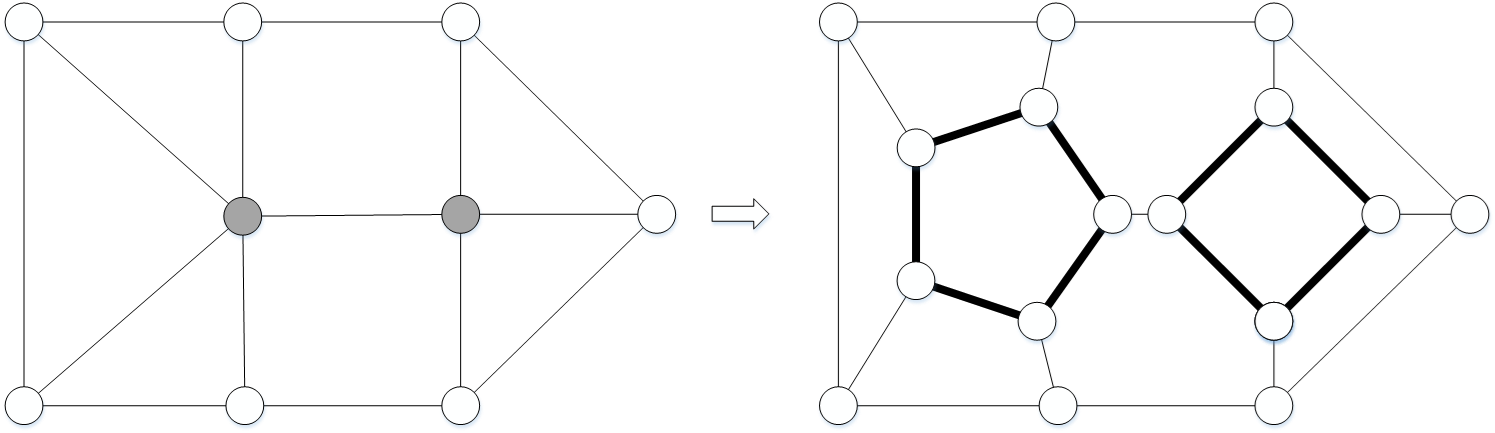}
    \caption{
    Demonstration of building an equivalent model with maximum degree
    $\Delta \leq 3$ via `expanding' vertices (in grey).
    In the new model, one can introduce edge factor $\psi_{u,v}$ between the duplicated vertices $u,v$ (in bold) such that
    $\psi_{u,v}(x_u,x_v) =1$ if $x_u=x_v$ and  $\psi_{u,v}(x_u,x_v) =0$ otherwise.}
    \label{fig:make3regular}
    \vspace{-0.1in}
\end{figure}

\section{Proof of Theorem \ref{thm:sample2regular}}
First, note that the MC induced by the worm algorithm 
converges to the following stationary distribution
\begin{equation*}
    \pi_{\text{WA}}(F) \propto \Psi(F)\prod_{e\in F}w(e),
\end{equation*}
where 
\begin{align*}
    \Psi(F) = 
    \begin{cases}
    n, \qquad &\forall F \in \mathcal{L}_{\text{2-Loop}},\\
    2, \qquad &\forall F \in \mathcal{L}_{\text{2-Odd}}.
    \end{cases}
\end{align*}
We first prove its polynomial mixing,
i.e. it produces a sample from a 
distribution with the 
desired total variation distance 
from $\pi_{\text{WA}}$ in a polynomial number of iterations.
\begin{lemma}\label{lem:mixing2regular}
Given any $\delta >0$ and any $F_{0}\in \mathcal{L}_{\text{2-Loop}}\cup\mathcal{L}_{\text{2-Loop}}$, 
choose 
\begin{align*}
T_{\text{mix}} \geq w(F_{0})^{-1} + (m-n+1)\log2 + 12\Delta mn^{4}\log \delta^{-1},
\end{align*} 
and let $\pi^{t}_{\text{WA}}(\cdot)$ denote the resulting distribution of 
after updating $t$ times by the worm algorithm with initial state $F_{0}$. 
Then, it follows that 
\begin{align*}
    \frac{1}{2}\sum_{F \in \mathcal{L}_{\text{2-Loop}}\cup\mathcal{L}_{\text{2-Loop}}}
    \bigg|
    \pi^{T_{\text{mix}}}_{\text{WA}}(F)
    - \pi_{\text{WA}}(F)
    \bigg|
    \leq \delta,
\end{align*}
namely, 
the mixing time of the MC is bounded above by $T_{\text{mix}}$. 
\end{lemma}
The proof of the above lemma is given in Section \ref{sec:pf:lem:mixing2regular}.
Collevecchio et al. \cite{C13} recently proved that the worm algorithm mixes in polynomial time when 
the weights are uniform, i.e., equal. 
We extend the result to our case of non-uniform weights. 
The proof is based on the method of {\it canonical path}, which views the state space as a graph and constructs a path between every pair of states having certain amount of 
flow defined by $\pi_{\text{WA}}$.
From Lemma \ref{lem:mixing2regular} with parameters 
\begin{align*}
    &N\leq 1.2n\log(3\delta^{-1}), \quad  T\leq(m-n+1)\log 2 + 4\Delta mn^{4}\log(3n\delta^{-1}),\quad \mbox{and} \quad
    &F_{0}\leftarrow \emptyset, 
\end{align*}
we obtain that the total variation distance between $\pi_{\text{WA}}$ and 
the distribution of updated states in line 4 of {Algorithm \ref{alg:sample2regular}} 
is at most $\frac{\delta}{3n}$. 
Next, we prove that the probability of acceptance in 
line $6$ of {Algorithm \ref{alg:sample2regular}} is sufficiently large. 
\begin{lemma}\label{lem:sample2regular}
The probability of sampling a $2$-regular loop from distribution $\pi_{WA}$ is 
bounded below by
$n^{-1}$, i.e. 
$
\pi_{\text{WA}}(\mathcal{L}_{\text{2-Loop}})
\geq\frac{1}{n}.
$
\end{lemma} 
The proof of the above lemma is given in Section \ref{sec:pf:lem:sample2regular}.
The proof relies on the fact that the size of $\mathcal{L}_{\text{2-Loop}}$ is 
bounded by a polynomial of the size of $\mathcal{L}_{\text{2-Odd}}$.

Now we are ready to complete the proof of Theorem \ref{thm:sample2regular}. 
Let $\widehat{\pi}_{\text{2-Loop}}$ denote the distribution of $2$-regular loops  
from line 6 of {Algorithm \ref{alg:sample2regular}} under parameters as in Theorem \ref{thm:sample2regular}. 
We say {Algorithm \ref{alg:sample2regular}} fails if it outputs
$F=\emptyset$ from line 9. 
Choose a set of $2$-regular loops 
$\widehat{\mathcal{L}}_{\text{2-Loop}}:= \{F\in \mathcal{L}_{\text{2-Loop}}: \widehat \pi_{\text{2-Loop}}(F) > \pi_{\text{2-Loop}}(F)\}.$
Then the total variation distance between $\pi_{\text{2-Loop}}$ and 
$\widehat{\pi}_{\text{2-Loop}}$
can be expressed as: 
\begin{align*}
\frac{1}{2}\sum_{F \in \mathcal{L}_{\text{2-Loop}}}|\widehat{\pi}_{\text{2-Loop}}(F) - \pi_{\text{2-Loop}}(F)| = 
\widehat{\pi}_{\text{2-Loop}}(\widehat{\mathcal{L}}_{\text{2-Loop}}) - \pi_{\text{2-Loop}}(\widehat{\mathcal{L}}_{\text{2-Loop}}).
\end{align*}
By applying Lemma \ref{lem:mixing2regular} and Lemma \ref{lem:sample2regular}, we obtain the following
under parameters as in Theorem \ref{thm:sample2regular}:
\begin{align*}
&\widehat{\pi}_{\text{2-Loop}}(\widehat{\mathcal{L}}_{\text{2-Loop}}) - \pi_{\text{2-Loop}}(\widehat{\mathcal{L}}_{\text{2-Loop}})\\
&\qquad \stackrel{(a)}{\geq}~ \frac{\widehat{\pi}_{\text{WA}}(\widehat{\mathcal{L}}_{\text{2-Loop}})}{\widehat{\pi}_{\text{WA}}(\mathcal{L}_{\text{2-Loop}})} - (1-\widehat{\pi}_{\text{WA}}(\mathcal{L}_{\text{2-Loop}}))^{N} - \pi_{\text{2-Loop}}(\widehat{\mathcal{L}}_{\text{2-Loop}}) \\
&\qquad\stackrel{(b)}{\geq}~ \frac{\pi_{\text{WA}}(\widehat{\mathcal{L}}_{\text{2-Loop}})+\frac{\delta}{3n}}{\pi_{\text{WA}}(\mathcal{L}_{\text{2-Loop}})-\frac{\delta}{3n}} - (1-\pi_{\text{WA}}(\mathcal{L}_{\text{2-Loop}})-\frac{\delta}{3n})^{N} - \pi_{\text{2-Loop}}(\widehat{\mathcal{L}}_{\text{2-Loop}}) \\
&\qquad\stackrel{(c)}{\geq}~
- \frac{2\delta}{3n\,\pi_{\text{WA}}(\mathcal{L}_{\text{2-Loop}})} - e^{-(\pi_{\text{WA}}(\mathcal{L}_{\text{2-Loop}}) + \frac{\delta}{3n})N}\\
&\qquad \stackrel{(d)}{\geq}~ 
- \frac{2\delta}{3} - \frac{\delta}{3} =  - \delta.
\end{align*}
In the above, (a) comes from the fact that 
a sample from line 6 of {Algorithm \ref{alg:sample2regular}} 
follows the distribution $\frac{\widehat{\pi}_{\text{WA}}(\widehat{\mathcal{L}}_{\text{2-Loop}})}{\widehat{\pi}_{\text{WA}}(\mathcal{L}_{\text{2-Loop}})}$ and the failure probability of {Algorithm \ref{alg:sample2regular}} is  $(1-\widehat{\pi}_{\text{WA}}(\mathcal{L}_{\text{2-Loop}}))^{N}$.
For (b),
we use the variation distance between $\widehat{\pi}_{\text{WA}}$ and $\pi_{\text{WA}}$ due to Lemma \ref{lem:mixing2regular}
and parameters as in Theorem \ref{thm:sample2regular}, i.e.,
\begin{equation*}
|\widehat{\pi}_{\text{WA}}(S) - \pi_{\text{WA}}(S)| \leq \frac{\delta}{3n}\qquad \forall~S\subseteq \mathcal{L}_{\text{2-Loop}}\cup\mathcal{L}_{\text{2-Odd}}.
\end{equation*}
For (c), we use $(1-x) \leq e^{-x}$ for any $x\geq 0$ and
(d) follows from Lemma \ref{lem:sample2regular}
and $N \leq n\ln(3\delta^{-1})$.
The converse $\widehat{\pi}_{\text{2-Loop}}(\widehat{\mathcal{L}}_{\text{2-Loop}}) - \pi_{\text{2-Loop}}(\widehat{\mathcal{L}}_{\text{2-Loop}}) \leq \delta$ can be done similarly
by considering the complementary set $\mathcal{L}_{\text{2-Loop}} \backslash \widehat{\mathcal{L}}_{\text{2-Loop}}$.
This completes the proof of Theorem \ref{thm:sample2regular}. 

\subsection{Proof of Lemma \ref{lem:mixing2regular}}\label{sec:pf:lem:mixing2regular} 
First, let $P_{\text{WA}}$ denote the transition matrix of MC induced by the worm algorithm in Section \ref{sec:sample2regular}. 
Then we are able to define the corresponding transition graph $\mathcal{G}_{\text{WA}} = (\mathcal{L}_{\text{2-Loop}} \cup \mathcal{L}_{\text{2-Odd}}, \mathcal{E}_{\text{WA}})$, 
where each vertex is a state of the MC, and edges are defined on 
state pairs with nonzero transition probability, i.e.
\begin{equation*}
\mathcal{E}_{\text{WA}} = \{(A, A^\prime): (A, A^\prime)\in (\mathcal{L}_{\text{2-Loop}} \cup \mathcal{L}_{\text{2-Odd}})\times (\mathcal{L}_{\text{2-Loop}} \cup \mathcal{L}_{\text{2-Odd}}), P_{\pi_{\text{WA}}}(A,A^\prime) > 0\}.
\end{equation*}
Our proof makes use of the following result proved in \cite{C13}. 
\begin{theorem}[Schweinsberg 2002 \cite{C13}] 
Consider an irreducible and lazy MC, 
with finite state space $\Omega$, transition matrix $P$ 
and transition graph $\mathcal{G}_{P}$, 
which is reversible with respect to the distribution $\pi$. 
Let $\mathcal{O} \subseteq \Omega$ be nonempty, and 
for each pair $(I,J) \in \Omega \times \mathcal{O}$, specify a path 
$\gamma_{I,J}$ in $\mathcal{G}_{P}$ from $I$ to $J$. 
Let 
\begin{equation*}
\Gamma = \{\gamma_{I,J} : (I,J) \in \Omega \times \mathcal{O}\}
\end{equation*}
denote the collection of all such paths, and let $L(\Gamma)$ be the length of longest 
path in $\Gamma$. 
For any transition $T\in \mathcal{E}_{P}$, let 
\begin{equation*}
\mathcal{H}_{T} = \{(I,F)\in \Omega \times \mathcal{O}: T \in \gamma_{I,J}\}.
\end{equation*}
Then 
\begin{equation*}
\tau_{A}(\delta) \leq \left[\log\left(\frac{1}{\pi(A) + \log \left(\frac{1}{\delta}\right)}\right)\right] 4L(T)\Phi(\Gamma)
\end{equation*}
where 
\begin{equation*}
\Phi(\Gamma) = \max_{(A,A^\prime) \in \mathcal{E}_{P}}
\left\{
\sum_{I,J \in \mathcal{H}_{(A,A^\prime)}} \frac{\pi (I)\pi(J)}{\pi(\mathcal{O})\pi(A)P(A,A^\prime)}
\right\}.
\end{equation*}
\end{theorem}

To this end, we choose $\mathcal{O} = \mathcal{L}_{\text{2-Loop}}$ and 
we show that there exists a choice of paths $\Gamma = \{\gamma_{I,J}: (I,J) \in (\mathcal{L}_{\text{2-Loop}}\cup \mathcal{L}_{\text{2-Loop}}) \times \mathcal{L}_{\text{2-Odd}}\}$ such that
\begin{equation*} 
\Phi(\Gamma) \leq \Delta n^{4}, \qquad L(\Gamma) \leq m.
\end{equation*}
Then we obtain the statement in Lemma \ref{lem:mixing2regular} 
immediately.

We begin by specifying $\Gamma$, and then proceed to the bound of 
$\Phi(\Gamma)$. 
To this end, we fix an $[n]$-valued vertex labeling of $\mathcal{G}_{\text{WA}}$. 
The labeling induces a lexicographical total order of the edges, 
which in turn induces a lexicographical total order on the 
set of all subgraphs of $\mathcal{G}_{\text{WA}}$.
In order for the state $I\in \mathcal{L}_{\text{2-Loop}}\cup \mathcal{L}_{\text{2-Odd}}$ 
transit to the $J\in \mathcal{L}_{\text{2-Loop}}$, 
it suffices that it updates, 
precisely once, those edges in $I\oplus J$. 
In order to describe such path,
we first prove that there exist a injection 
from $I\oplus J$ 
to some unique disjoint partition $I\oplus J = \cup^{k}_{i=0}C_{i}$, where 
$C_{0}$ is either a path or a cycle and 
$C_{1}, \cdots, C_{k}$ are cycles. 
Observe that since $J \in \mathcal{L}_{\text{2-Loop}}$, 
applying symmetric difference with $J$ does not change the parity of 
degrees of the vertices and 
$I \oplus J \in \mathcal{L}_{\text{2-Loop}}\cup \mathcal{L}_{\text{2-Odd}}$. 
First consider the case when $I\oplus J \in \mathcal{L}_{\text{2-Odd}}$. 
Then there exist a path 
between two odd-degree vertices in $I\oplus J$, since 
the sum of degrees over all vertices in a component is even.
Among such paths, we pick $C_{0}$ as the path 
with the highest order according to the $[n]$-valued vertex labeling. 
Now observe that $I\oplus J \backslash C_{0} \in \mathcal{L}_{\text{2-Loop}}$ is 
Eulerian, which can be decomposed into disjoint set of cycles. 
We are able to choose a $C_{1}, \cdots, C_{k}$ uniquely 
by recursively excluding a cycle with the highest order, 
i.e. we pick $C_{1}$ as a cycle with highest order from 
$I\oplus J\backslash C_{0}$, 
then pick $C_{2}$ from $I\oplus J \backslash C_{0}\backslash C_{1}$ 
with the highes order, and so on.
For the case when $I \in \mathcal{L}_{\text{2-Loop}}$, 
$I\oplus J\in \mathcal{L}_{\text{2-Loop}}$ is Eulerian and we can 
apply similar logic to obtain the unique decomposition into disjoint cycles.

Now we are ready to describe $\gamma_{I,J}$, which updates 
the edges in $I\oplus J$ from $C_{0}$ to $C_{k}$ in order. 
If $C_{0}$ is a path, 
pick an endpoint with higher order of label and update the 
edges in the paths by {it unwinding} the edges 
along the path until other endpoint is met. 
In the case of cycles, pick a vertex 
with highest order of label and 
unwind the edges by a fixed orientation. 
Note that during the update of cycles, 
the number of odd-degree vertices are at most 2, 
so the
intermediate states are stil 
in $\mathcal{L}_{\text{2-Loop}} \cup L_{\text{2-Odd}}$. 
As a result, we have constructed a path $\gamma_{I, F}$ 
for each $I \in \mathcal{L}_{\text{2-Loop}}\cup\mathcal{L}_{\text{2-Odd}}$ 
and $J\in \mathcal{L}_{\text{2-Loop}}$ where 
each edge correspond to an update on $I\oplus J$ and 
$|\gamma_{I,F}| = |I\oplus J| \leq m$.

Next, we bound the corresponding $\Phi(\Gamma)$. 
First let 
$\mathcal{L}_{\text{4-Odd}}$ denote the 
set of subgraphs with exactly $4$ odd-degree vertices. We define a mapping 
$\eta_{T}: \mathcal{H}_{T}\rightarrow \mathcal{L}_{\text{2-Loop}}\cup \mathcal{L}_{\text{2-Odd}}\cup \mathcal{L}_{\text{4-Odd}}$ by the following:
\begin{equation*}
\eta_{T}(I,J) := I\oplus F\oplus (A\cup e),
\end{equation*}
where $T = (A, A\oplus e)$. 
Observe that $\eta_{T}(I,J)$ agrees with $I$ on the components 
that have already been processed, 
and with $J$ on the components that have not.
We prove that $\eta_{T}$ is an injection 
by reconstructing $I$ and $J$ from $\eta_{T}(I,J)$ given $T= (A, A\oplus e)$. 
To this end, 
observe that $I\oplus F = \eta_{T}(I,F) \oplus (A \cup e)$ is 
uniquely decided from $\eta_{T}(I,F)$ and $(A\cup e)$. 
Then given $I\oplus F$, we are able to infer the 
decomposition $C_{0}, C_{1}, \cdots, C_{k}$ of $I\oplus J$ 
by the rules defined previously. Moreover the updated edge $e$ implies the current set $C_{i}$ being updated. 
Therefore we can infer the processed part of $I\oplus J$. 
Then we can recover $J$ by beginning in $A$ and 
unwinding the remaining edges in $I\oplus J$ that was not processed yet. 
Then we recover $I$ via $I = \eta_{T}(I,J)\oplus (A\cup e)\oplus J$ 
and therefore $\eta_{T}$ is injective.

Next, 
we define a metric $w_{\text{WA}}$ 
such that given an edge set $F$,  
\begin{equation*}
    w_{\text{WA}}(F):= \prod_{e\in F}|w(e)|.
\end{equation*} 
We complete the proof by showing that for any 
$T = (A, A^\prime) \in \mathcal{E}$, the following inequality holds: 
\begin{align*}
\Phi(\Gamma) \stackrel{(a)}{\leq}
\sum_{I,J \in \mathcal{H}_{T}}
\frac{1}{\pi(\mathcal{L}_{\text{2-Loop}})} \frac{\pi(I)\pi(J)}{\pi(A)P(A,A^\prime)} 
\stackrel{(b)}{\leq} 
\sum_{I,J \in \mathcal{H}_{T}}
\frac{2\Delta}{w_{\text{WA}}(\mathcal{L}_{\text{2-Loop}})} \Psi(I) w_{\text{WA}}(\eta_{T}(I,J)) 
\stackrel{(c)}{\leq}\Delta n^{4}.
\end{align*}
First, (a) holds by definition of $\Phi$. 
We prove (b) by the following chain of inequality:
\begin{align*}
\frac{1}{\pi(\mathcal{L}_{\text{2-Loop}})} \frac{\pi(I)\pi(J)}{\pi(A)P(A,A^\prime)} 
&=\frac{1}{nw_{\text{WA}}(\mathcal{L}_{\text{2-Loop}})} \frac{\Psi(I)w_{\text{WA}}(I)nw_{\text{WA}}(J)}{\Psi(A)w_{\text{WA}}(A)P_{\text{WA}}(A,A^\prime)}\\
&\stackrel{(1)}{\leq} \frac{1}{w_{\text{WA}}(\mathcal{L}_{\text{2-Loop}})} \Psi(I)w_{\text{WA}}(I)w_{\text{WA}}(J) \frac{2\Delta}{w_{\text{WA}}(A\cup e)} \\
&\stackrel{(2)}{=} \frac{2\Delta}{w_{\text{WA}}(\mathcal{L}_{\text{2-Loop}})}\Psi(I) w_{\text{WA}}(\eta_{T}(I,F)).
\end{align*}
In the above, (1) comes from the definition of the transition probability 
and (2) comes from the definition of function $w_{\text{WA}}$.
Finally, we prove (c). First, we have
\begin{align*}
\Psi(\Gamma) 
&\leq \sum_{(I,J)\in \mathcal{H}_{T}}\frac{2\Delta}{w_{\text{WA}}(\mathcal{L}_{\text{2-Loop}})}\Psi(I) w_{\text{WA}}(\eta_{T}(I,F))\\
&\leq \sum_{(I,J)\in \mathcal{H}_{T}} 
\frac{2\Delta}{w_{\text{WA}}(\mathcal{L}_{\text{2-Loop}})}[w_{\text{WA}}(\mathcal{L}_{\text{2-Loop}}\cup \mathcal{L}_{\text{2-Odd}}) + 2w_{\text{WA}}(\mathcal{L}_{\text{2-Loop}}\cup \mathcal{L}_{\text{2-Odd}} \cup \mathcal{L}_{\text{4-Odd}})]\\
&= 2\Delta \left[(n+2) + (n+2) + \frac{w_{\text{WA}}(\mathcal{L}_{\text{2-Odd}})}{w_{\text{WA}}(\mathcal{L}_{\text{2-Loop}})} + 2\frac{w_{\text{WA}}(\mathcal{L}_{\text{4-Odd}})}{w_{\text{WA}}(\mathcal{L}_{\text{2-Loop}})}\right],
\end{align*}
since $\eta_{T}(I,J)$ is an injection on $\mathcal{L}_{\text{2-Loop}}\cup \mathcal{L}_{\text{2-Odd}}\cup \mathcal{L}_{\text{4-Odd}}$, 
and the set $\mathcal{L}_{\text{2-Loop}}, \mathcal{L}_{\text{2-Odd}}, \mathcal{L}_{\text{4-Odd}}$ are disjoint. Now we prove 
\begin{equation*}
\frac{w_{\text{WA}}(\mathcal{L}_{\text{2-Odd}})}{w_{\text{WA}}(\mathcal{L}_{\text{2-Loop}})} \leq \binom{n}{2} \qquad 
\frac{w_{\text{WA}}(\mathcal{L}_{\text{4-Odd}})}{w_{\text{WA}}(\mathcal{L}_{\text{2-Loop}})} \leq \binom{n}{4},
\end{equation*}
which completes the proof of Lemma \ref{lem:mixing2regular} since 
$
(n+2) + (n+2) + \binom{n}{2} + 2\binom{n}{4}
\leq \frac{n^{4}}{2}.
$
To this end, we let $\mathcal{L}_{\text{Odd}}(W)$
denote the set of generalized loops having 
$W$ as the set of odd degree vertices. 
Now observe the following inequality:
\begin{align*}
\sum_{F\in \mathcal{L}_{\text{Odd}}(W)} w_{\text{WA}}(F) 
&\stackrel{(a)}{=}~ \frac{1}{2^{n}}\sum_{F\in \mathcal{L}} \prod_{e\in F} |w(e)| \prod_{s\in V\backslash W} (1 + (-1)^{d_{F}(v)})
\prod_{s\in W}(1 + (-1)^{d_{F}(v)+1})\\
&= \frac{1}{2^{n}}\sum_{\sigma\in\{-1,1\}^V}\sum_{F\in \mathcal{L}} \prod_{e\in F} |w(e)| 
\prod_{s\in V} \sigma_{v}^{d_{F}(v)} \prod_{v\in W} \sigma_{v}\\
&=~ \sum_{\sigma \in \{-1, +1\}^{V}}\prod_{e=(u,v)\in E}(1+ |w(e)|\sigma_{u}\sigma_{v})\prod_{v\in W} \sigma_{v}\\
&\stackrel{(b)}{\geq}~ \sum_{\sigma \in \{-1, +1\}^{V}}\prod_{e=(u,v)\in E}(1+ |w(e)|\sigma_{u}\sigma_{v})\\
&\stackrel{(c)}{=}~ \sum_{F\in \mathcal{L}_{\mathcal{L}_{\text{2-Loop}}}} w_{\text{WA}}(F).
\end{align*}
In the above, (a) comes from the fact that 
$1 + (-1)^{d_{v}(F)} = 2$ if $d_{v}(F)$ is even and $0$ otherwise, so only the terms corresponding to $2$-regular loop becomes non-zero. 
For (b), the inequality comes from the fact that $1+ |w(e)|\sigma_{u}\sigma_{v}\geq 0$ and $\sigma_{v} \leq 1$.
For (c), the equality is from the fact that $\mathcal{L}_{\text{2-Loop}} =\mathcal{L}_{\text{Odd}}(\emptyset).$
Therefore we have $\sum_{F\in L(\emptyset)}|w(F)| \geq \sum_{F\in L(W)}|w(F)|$, leading to 
\begin{align*}
\frac{w_{\text{WA}}(\mathcal{L}_{\text{2-Odd}})}{w_{\text{WA}}(\mathcal{L}_{\text{2-Loop}})}
&= 
\frac{\sum_{W\subseteq V, |W|=2}\sum_{F \in \mathcal{L}_{\text{Odd}}(W)}|w_{\text{WA}}(F)|}{
w_{\text{WA}}(\mathcal{L}_{\text{2-Loop}})} 
\leq \binom{n}{2},
\end{align*}
and the case for $\mathcal{L}_{\text{4-Odd}}$ is done similarly. This completes the proof of Lemma \ref{lem:mixing2regular}.

\subsection{Proof of Lemma \ref{lem:sample2regular}}\label{sec:pf:lem:sample2regular}
Given $W\subseteq V$, 
we let $\mathcal{L}_{\text{Odd}}(W)$
denote the set of generalized loops having 
$W$ as the set of odd degree vertices. 
where $\mbox{Odd}(F)$ is the set of odd-degree vertices in $F$.
Now observe the following inequality:
\begin{align*}
\sum_{F\in \mathcal{L}_{\text{Odd}}(W)} w_{\text{WA}}(F) 
&\stackrel{(a)}{=}~ \frac{1}{2^{n}}\sum_{F\in \mathcal{L}} \prod_{e\in F} |w(e)| \prod_{s\in V\backslash W} (1 + (-1)^{d_{F}(v)})
\prod_{s\in W}(1 + (-1)^{d_{F}(v)+1})\\
&= \frac{1}{2^{n}}\sum_{\sigma\in\{-1,1\}^V}\sum_{F\in \mathcal{L}} \prod_{e\in F} |w(e)| 
\prod_{s\in V} \sigma_{v}^{d_{F}(v)} \prod_{v\in W} \sigma_{v}\\
&=~ \sum_{\sigma \in \{-1, +1\}^{V}}\prod_{e=(u,v)\in E}(1+ |w(e)|\sigma_{u}\sigma_{v})\prod_{v\in W} \sigma_{v}\\
&\stackrel{(b)}{\geq}~ \sum_{\sigma \in \{-1, +1\}^{V}}\prod_{e=(u,v)\in E}(1+ |w(e)|\sigma_{u}\sigma_{v})\\
&\stackrel{(c)}{=}~ \sum_{F\in \mathcal{L}_{\mathcal{L}_{\text{2-Loop}}}} w_{\text{WA}}(F).
\end{align*}
In the above, (a) comes from the fact that 
$1 + (-1)^{d_{v}(F)} = 2$ if $d_{v}(F)$ is even and $0$ otherwise, so only the terms corresponding to $2$-regular loop becomes non-zero. 
For (b), the inequality comes from the fact that $1+ |w(e)|\sigma_{u}\sigma_{v}\geq 0$ and $\sigma_{v} \leq 1$.
For (c), the equality is from the fact that $\mathcal{L}_{\text{2-Loop}} =\mathcal{L}_{\text{Odd}}(\emptyset).$
Therefore we have $\sum_{F\in L(\emptyset)}|w(F)| \geq \sum_{F\in L(W)}|w(F)|$, leading to 
\begin{align*}
\frac{\sum_{F \in \mathcal{L}_{\text{2-Loop}}}\pi_{\text{WA}}(F)}{\sum_{F \in \mathcal{L}_{\text{2-Loop}}\cup \mathcal{L}_{\text{2-Odd}}}\pi_{\text{WA}}(F)} 
&= \frac{n\sum_{F \in \mathcal{L}_{\text{2-Loop}}}|w_{\text{WA}}(F)|}{n\sum_{F \in \mathcal{L}_{\text{2-Loop}}}|w_{\text{WA}}(F)| + \sum_{W_{\text{WA}}\subseteq V, |W|=2}\sum_{F \in \mathcal{L}_{\text{Odd}}(W)}|w_{\text{WA}}(F)|} \\
&\geq \frac{n}{n+2\binom{n}{2}} = \frac{1}{n},
\end{align*}
which completes the proof of Lemma \ref{lem:sample2regular}.

\section{Proof of Theorem \ref{thm:estimatepartition2regular}}
First, we quantify how much samples from {Algorithm \ref{alg:sample2regular}} 
are necessary for 
estimating some non-negative real valued function $f$ on $\mathcal{L}_{\text{2-Loop}}$. 
To this, we state the following lemma which is a straightforward application
of the known result in \cite{C3}.
\begin{lemma}
\label{lem:samplenum2regular}
Let $f$ be a non-negative real-valued function defined on $\mathcal{L}_{\text{2-Loop}}$
and bounded above by $f_{\max}\geq 0$.
Given $0<\xi\leq 1$ and $0<\eta \leq 1/2$, 
choose 
\begin{align*}
s &\geq\frac{504\xi^{-2}\lceil\log \eta^{-1}\rceil f_{\max}}{\mathbb{E}_{\pi_{\text{2-Loop}}}[f]}\qquad
N \geq 1.2n \log\frac{24f_{\max}}{\xi\mathbb{E}_{\pi_{\text{2-Loop}}}[f]}, \\
T &\geq (m-n+1)\log 2 + 4\Delta mn^{4} \log\frac{8f_{\max}}{\xi\mathbb{E}_{\pi_{\text{2-Loop}}}[f]},
\end{align*}
and generate $2$-regular loops $F_{1}, F_{2}, \cdots F_{s}$ using 
{Algorithm \ref{alg:sample2regular}} with inputs $N$ and $T$.
Then, it follows that 
\begin{align*}
    P\bigg[
    \frac{ |\frac1s\sum_{i}|w(F_{i})| - \mathbb{E}_{\pi_{\text{2-Loop}}}(f)|}
    {\mathbb{E}_{\pi_{\text{2-Loop}}}(f)} 
    \leq \xi\bigg] \leq 1-\eta.
\end{align*}
namely, samples of {Algorithm \ref{alg:sample2regular}} estimates $\mathbb{E}_{\pi_{\text{2-Loop}}}(f)$ 
within approximation ratio $1\pm\xi$ with probability at least $1-\eta$.
\end{lemma} 
First, recall that during each stage of simulated annealing, 
we approximate the expectation of the function $w(F)^{1/n}$
with respect to the distribution $\pi_{\text{2-Loop}}(\beta)$, 
i.e., 
\begin{equation*}\mathbb{E}_{\pi_{\text{2-Loop}}(\beta)}\left[|w(F)|^{1/n}\right] = Z^{\dagger}_{\text{2-Loop}}(\beta_{i+1})/Z^{\dagger}_{\text{2-Loop}}(\beta_{i}).\end{equation*} 
Hence, to apply Lemma \ref{lem:samplenum2regular},
we bound $\max_F{|w(F)|^{1/n}}$ and $\mathbb{E}_{\pi_{\text{2-Loop}}(\beta)}\left[|w(F)|^{1/n}\right]$ 
as follows:
\begin{equation*}
|w(F)|^{1/n}
\leq 1
\qquad\qquad
\mathbb{E}_{\pi_{\text{2-Loop}}(\beta)}\left[|w(F)|^{1/n}\right]
\geq~ w_{\min},
\end{equation*}
where the first inequality is due to $w(e) \leq 1$ for any $e\in E$ and 
the second one is from $|F|\leq n$ for any $2$-regular loop $F$.
Thus, from Lemma \ref{lem:samplenum2regular} 
with parameters
\begin{align*}
    &s\geq 18144 n^{2}\varepsilon^{-2}w^{-1}_{\min}\lceil\log (6n\nu^{-1})\rceil,\qquad
&&N \geq  1.2n\log(144n \varepsilon^{-1}w_{\min}^{-1}),\\
&T \geq
(m-n+1)\log 2 + 4\Delta mn^{4}\log(48 n \varepsilon^{-1}w_{\min}^{-1}),
\end{align*}
on each stage, we obtain
\begin{equation*}
P\bigg[\frac{|H_{i} - Z^{\dagger}_{\text{2-Loop}}(\beta_{i+1})/Z^{\dagger}_{\text{2-Loop}}(\beta_{i})|}
{Z^{\dagger}_{\text{2-Loop}}(\beta_{i+1})/Z^{\dagger}_{\text{2-Loop}}(\beta_{i})}
\leq \frac{\varepsilon}{6n}\bigg] 
\geq 1-\frac{\nu}{6n}.
\end{equation*}
This implies that
the product $\prod_{i}H_{i}$ estimates 
$\frac{Z^{\dagger}_{\text{2-Loop}}}{2^{m-n+1}}$ 
within approximation ratio in
\begin{equation*}[((1-\varepsilon/6n)^{n}, (1+\varepsilon/6n)^{n}] \subseteq [1 - \varepsilon/3, 1 + \varepsilon/3]\end{equation*} 
with probability at least $(1-\nu/6n)^{n} \geq 1 - \nu/3$, i.e.,
\begin{equation*}
P\bigg[
\frac{|2^{m-n+1}\prod_{i}H_{i} - Z^{\dagger}_{\text{2-Loop}}|}
{Z^{\dagger}_{\text{2-Loop}}}
\leq \frac{\varepsilon}{3}
\bigg]
\geq 1 - \frac{\nu}{3}.
\end{equation*}

Next we define a non-negative real-valued random function $g$ on $\mathcal{L_{\text{2-Loop}}}$ as
\begin{equation*}
g(F) = 
\begin{cases}
1 \qquad \mbox{if $w(F) < 0$} \\
0 \qquad \mbox{otherwise}
\end{cases},
\end{equation*}
namely, 
$\mathbb{E}_{\pi_{\text{2-Loop}}}\left[g(F)\right] = P_{\pi_{\text{2-Loop}}}[w(F) < 0]$.
Since $\max_F {g(F)} = 1$,  
one can apply Lemma \ref{lem:samplenum2regular} with parameters
\begin{align*}
    &s \geq  18144\zeta(1-2\zeta)^{-2}
\varepsilon^{-2}\lceil\log (3\nu^{-1})\rceil,
\qquad\qquad\qquad
&&N \geq
1.2n\log (144\varepsilon^{-1}
(1-2\zeta)^{-1}),\\
&T \geq 
(m-n+1)\log 2 + 4\Delta mn^{4}\log(48\varepsilon^{-1}
(1-2\zeta)^{-1})
\end{align*}
and have
\begin{equation*}
P\bigg[
    \frac{|\kappa - P_{\pi_{\text{2-Loop}}}[w(F) < 0]|}
    {P_{\pi_{\text{2-Loop}}}[w(F) < 0]} \leq 
    \frac{(1-2P_{\pi_{\text{2-Loop}}}[w(F) < 0])\varepsilon}{6P_{\pi_{\text{2-Loop}}}[w(F) < 0]}
    \bigg]
    \geq 1-\frac{\nu}{3}, 
\end{equation*}
since $\zeta = P_{\pi_{\text{2-Loop}}}[w(F) < 0]$.
Furthermore, after some algebraic calculations, 
one can obtain 
\begin{equation*}
P \bigg[\frac{|(1-2\kappa) - (1- 2P_{\pi_{\text{2-Loop}}}[w(F) < 0])|}{1- 2P_{\pi_{\text{2-Loop}}}[w(F) < 0]}\leq \frac{\varepsilon}{3} \bigg]
\geq 1-\frac{\nu}{3}.
\end{equation*}
The rest of the proof is straightforward 
since we estimate $Z_{\text{2-Loop}} = (1-2P_{\pi_{\text{2-Loop}}}[w(F) < 0])Z^{\dagger}_{\text{2-Loop}}$ by 
$(1-2\kappa)2^{m-n+1}\prod_{i}H_{i}$, the approximation ratio is in 
$[(1-\varepsilon/3)^{2}, (1+\varepsilon/3)^{2}] \subseteq [1-\varepsilon, 1+ \varepsilon]$ 
with probability at least $(1-\nu/3)^{2} \geq 1-\nu$.

\section{Proof of Theorem \ref{thm:generalizedloopdecomp}}
Given $F \in \mathcal{L}$, 
we let the odd-degree vertices in $F$ (i.e., $d_{F}(\cdot)$ is odd) 
by $v_{1}, v_{2}, \cdots v_{2\ell}$ for some integer $\ell\geq 0$. 
Since we assume $G$ is connected, there exist a set of paths 
$P_{1},P_{2},\cdots P_{\ell}$ such that 
$P_{i}$ is a path from $v_{2i-1}$ to $v_{2i}$. 
Note that given any set of edges $D\subseteq E$,  
$D \oplus P_{i}$ changes the parities of $d_{D}(v_{2i-1}), d_{D}(v_{2i})$, while 
others remain same. 
Therefore, 
all degrees in $F \oplus P_{1} \oplus \cdots \oplus P_{\ell}$ become even.
Then, due to the definition of cycle basis,
there exist some $C_{1}, C_{2}, \cdots C_{k} \in \mathcal{C}$ 
such that 
\begin{equation*}
C_{1}\oplus C_{2} \cdots \oplus C_{k} = F \oplus P_{1} \oplus \cdots \oplus P_{\ell},
\end{equation*}
namely,
\begin{equation*}F=C_{1}\oplus C_{2} \cdots \oplus C_{k} \oplus P_{1} \oplus \cdots \oplus P_{\ell}.
\end{equation*}
This completes the proof of Theorem \ref{thm:generalizedloopdecomp}.


\begin{thebibliography}{9}

\bibitem{C1_1}
J. Pearl,  
``Probabilistic reasoning in intelligent systems: networks of plausible inference,''
\emph{Morgan Kaufmann}, 2014.

\bibitem{C1_2}
R. G. Gallager,  
``Low-density parity-check codes,''
\emph{Information Theory, IRE Transactions} 8(1): 21-28, 1962.

\bibitem{C1_3}
R. F. Kschischang, and J. F. Brendan, 
``Iterative decoding of compound codes by probability propagation in graphical models,''
\emph{Selected Areas in Communications, IEEE Journal} 16(2): 219-230, 1998.

\bibitem{C1_4}
M. I. Jordan, ed. ``Learning in graphical models,'' {\it Springer Science \& Business Media} 89, 1998.

\bibitem{C1_5}
R.J. Baxter, ``Exactly solved models in statistical mechanics,'' {\it Courier Corporation}, 2007.


\bibitem{C1_6}
W.T. Freeman, C.P. Egon, and T.C. Owen,
``Learning low-level vision.''
\emph{International journal of computer vision} 40(1): 25-47, 2000.

\bibitem{C2}
V. Chandrasekaran, S. Nathan, and H. Prahladh, ``Complexity of Inference in Graphical Models,'' 
\emph{Association for Uncertainty and Artificial Intelligence}, 2008

\bibitem{C3}
M. Jerrum, and A. Sinclair, ``Polynomial-time approximation algorithms for the Ising model,'' 
\emph{SIAM Journal on computing} 22(5): 1087-1116, 1993.

\bibitem{C5}
C. Andrieu, N. Freitas, A. Doucet, and M. I. Jordan,
``An introduction to MCMC for machine learning,''
\emph{Machine learning} 50(1-2), 5-43, 2003.

\bibitem{C6}
J. Pearl, ``Reverend Bayes on inference engines: A distributed hierarchical approach,'' 
\emph{Association for the Advancement of Artificial Intelligence}, 1982.


\bibitem{C8}
M. Chertkov, and V. Y. Chernyak, ``Loop series for discrete statistical models on graphs,'' 
\emph{Journal of Statistical Mechanics: Theory and Experiment} 2006(6): P06009, 2006.

\bibitem{C9}
M. Chertkov, V. Y. Chernyak, and R. Teodorescu, ``Belief propagation and loop series on planar graphs,'' 
\emph{Journal of Statistical Mechanics: Theory and Experiment} 2008(5): P05003, 2008.

\bibitem{C9-1}
V. Gomez, J. K. Hilbert, and M. Chertkov, 
``Approximate inference on planar graphs using Loop Calculus and Belief Propagation,'' 
\emph{The Journal of Machine Learning Research}, 11: 1273-1296, 2010.

\bibitem{C10}
P. W. Kasteleyn, ``The statistics of dimers on a lattice,'' 
\emph{Classic Papers in Combinatorics. Birkhäuser Boston}, 281-298, 2009.


\bibitem{C12}
N. Prokof'ev, and B. Svistunov, ``Worm algorithms for classical statistical models,'' 
\emph{Physical review letters} 87(16): 160601, 2001.


\bibitem{C11}
J.D. Horton, ``A polynomial-time algorithm to find the shortest cycle basis of a graph.'' {\it SIAM Journal on Computing} 16(2): 358-366, 1987.
APA	


\bibitem{C14}
H. A. Kramers, and G. H. Wannier, 
``Statistics of the two-dimensional ferromagnet. Part II,'' 
\emph{Physical Review} 60(3): 263, 1941.

\bibitem{C13}
A. Collevecchio, T. M. Garoni, T.Hyndman, and D. Tokarev, 
``The worm process for the Ising model is rapidly mixing,''
arXiv preprint arXiv:1509.03201, 2015.

\bibitem{SA}
S. Kirkpatrick, 
``Optimization by simulated annealing: Quantitative studies.''
{\it Journal of statistical physics} 34(5-6): 975-986, 1984.
\bibitem{C15}
R. Nicholas,
``The Bethe partition function of log-supermodular graphical models,'' 
\emph{Advances in Neural Information Processing Systems}. 2012.

\bibitem{BF}
J. Bang, J., and G. Z. Gutin. ``Digraphs: theory, algorithms and applications.'' {\it Springer Science \& Business Media}, 2008.

\bibitem{ConvBP1}
Y. W. Teh and M. Welling, 
``Belief optimization for binary networks: a stable alternative
to loopy belief propagation,'' 
\emph{Proceedings of the Eighteenth conference on Uncertainty in
artificial intelligence}, 493-500, 2001.

\bibitem{ConvBP2}
A. L. Yuille,
``CCCP algorithms to minimize the Bethe and Kikuchi free energies: Convergent
alternatives to belief propagation,'' 
\emph{Neural Computation}, 14(7): 1691-1722, 2002.

\bibitem{ConvBP3}
J. Shin, 
``The complexity of approximating a Bethe equilibrium,'' 
\emph{Information Theory, IEEE Transactions on}, 60(7): 3959-3969, 2014.

\bibitem{sign}
https://www.quora.com/Statistical-Mechanics-What-is-the-fermion-sign-problem

\bibitem{Independent}
Dyer, M., Frieze, A., and Jerrum, M. ``On counting independent sets in sparse graphs,'' {\it SIAM Journal on Computing} 31(5): 1527-1541, 2002.

\bibitem{appendthm}
J. Schweinsberg, ``An $O(n^{2})$ bound for the relaxation time of a Markov chain on cladograms.'' {\it Random Structures \& Algorithms} 20(1): 59-70, 2002.


\end{thebibliography}
\end{document}